\documentclass[letterpaper, 10 pt, conference]{ieeeconf}  

\IEEEoverridecommandlockouts                              

\usepackage{graphics,graphicx,float,subcaption,booktabs,xcolor,multirow,array,color,ifthen,tabu,colortbl,dblfloatfix,url,xparse,mathtools,patchcmd,amssymb,xspace,nicefrac,microtype,amsmath,amsfonts,bm,ragged2e,tikz,stackengine,etoolbox,xpatch,enumerate,xstring,setspace,tabularx,makecell,changepage,cuted,titlesec,wrapfig,tcolorbox,gensymb,algorithm,algpseudocode,makecell,cite}
\usepackage[bookmarksopen,bookmarksnumbered,
pdfpagemode=UseOutlines,
colorlinks=true,
linkcolor=red,
anchorcolor=blue,
citecolor=blue,
filecolor=blue,
menucolor=blue,
urlcolor=blue,
backref=false,pagebackref=false]{hyperref}
\usepackage[capitalise,noabbrev,nameinlink]{cleveref}
\usepackage[english]{babel}
\usepackage[font=small]{caption}
\usepackage[margin=1in]{geometry}
\usepackage{flafter} 
\newcommand{\website}{https://flash-aloha.github.io}
\newcommand{\method}{Flash ALOHA}

\titlespacing*{\section}{0pt}{0.4\baselineskip}{0\baselineskip}
\titlespacing*{\subsection}{0pt}{0.4\baselineskip}{0\baselineskip}
\titlespacing*{\paragraph}{0pt}{0.0\baselineskip}{0\baselineskip}

\title{\LARGE \bf
\method{}: Learning Fast and Precise Manipulation
}

\author{John Hua Yao \quad Qi Wu \quad Yihuai Gao \quad Chelsea Finn \quad Zipeng Fu \\
Stanford University\\
\href{\website}{Project Website}
}

\begin{document}
\twocolumn[{%
\renewcommand\twocolumn[1][]{#1}%
\maketitle
\begin{center}
    \vspace{-0.3in}
    \centering
    \captionsetup{type=figure}
    \includegraphics[width=\textwidth]{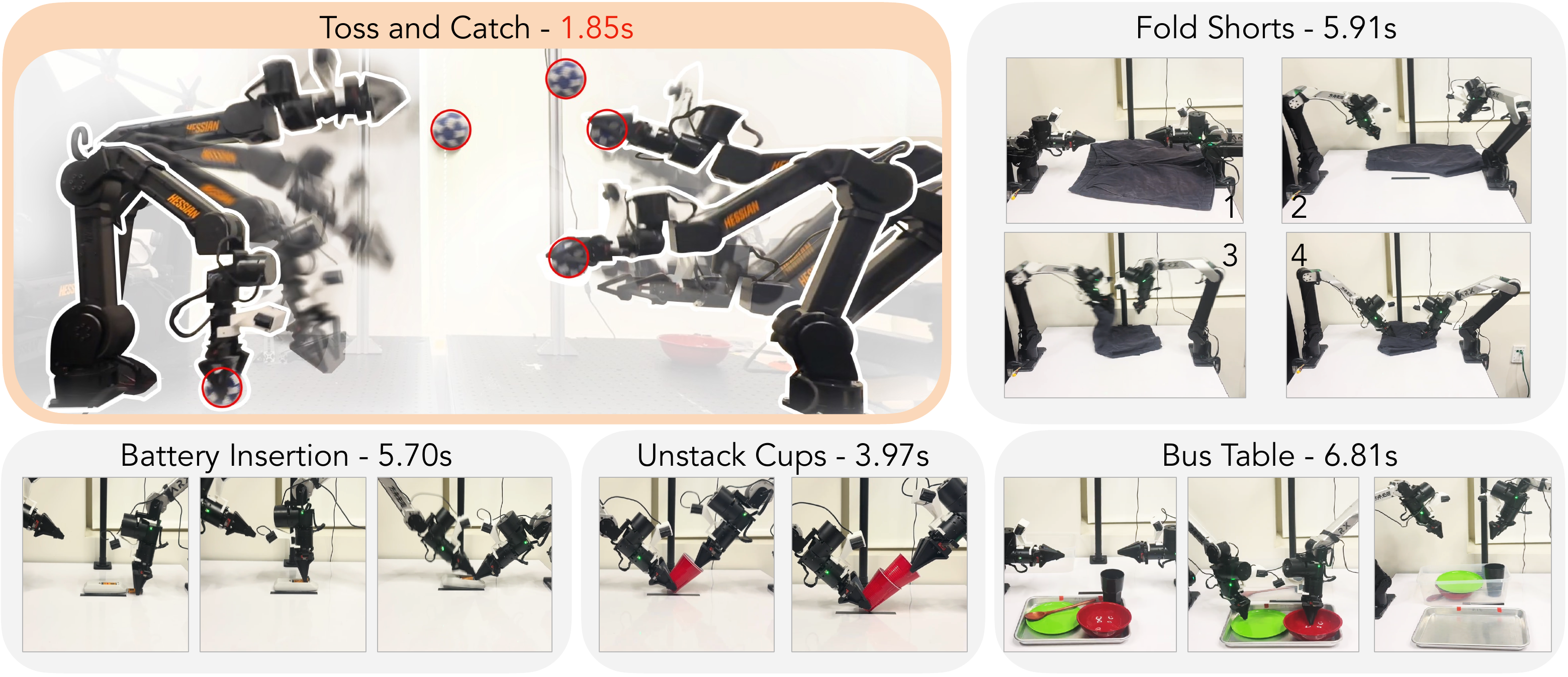}
    \captionof{figure}{\textbf{\textit{\method}}. We present a full-stack system for learning precise and high-speed manipulation policies, which can autonomously do bimanual toss and catch in under 2 seconds, bus a full dining set under 7, and fold a pair of shorts under 6.
    }
    \label{fig:teaser}
\end{center}%
}]
\thispagestyle{empty}
\pagestyle{empty}

\begin{abstract}
Learning from human demonstrations has enabled robots to acquire a wide range of manipulation skills, but learned policies typically execute far slower than ordinary humans. This speed gap is mainly due to lack of an interface for precisely collecting demonstration data at high speed, and the difficulty in training policies that can robustly execute high-speed motions. In this paper, we present \method{}, a practical system for learning fast and precise robotic manipulation. Our system uses kinesthetic teaching to intuitively collect accurate and near-human-speed demonstrations on a backdrivable bimanual platform, yielding natural and fast trajectories. Our learning pipeline consists of two components: a lightweight smoothing algorithm to address tracking instabilities produced by action chunking, and for visual alignment we mask out the operator visible in the camera views. Given 50 demonstrations for each task, \method{} autonomously completes bimanual tasks such as ball tossing and catching, folding shorts, battery insertion, and bussing tables at over 80\% success rates while close to human speed. 
\end{abstract}

\section{Introduction}

\begin{figure*}[t]
    \centering
    \begin{subfigure}[b]{0.4\textwidth}
        \centering
        \includegraphics[width=\linewidth]{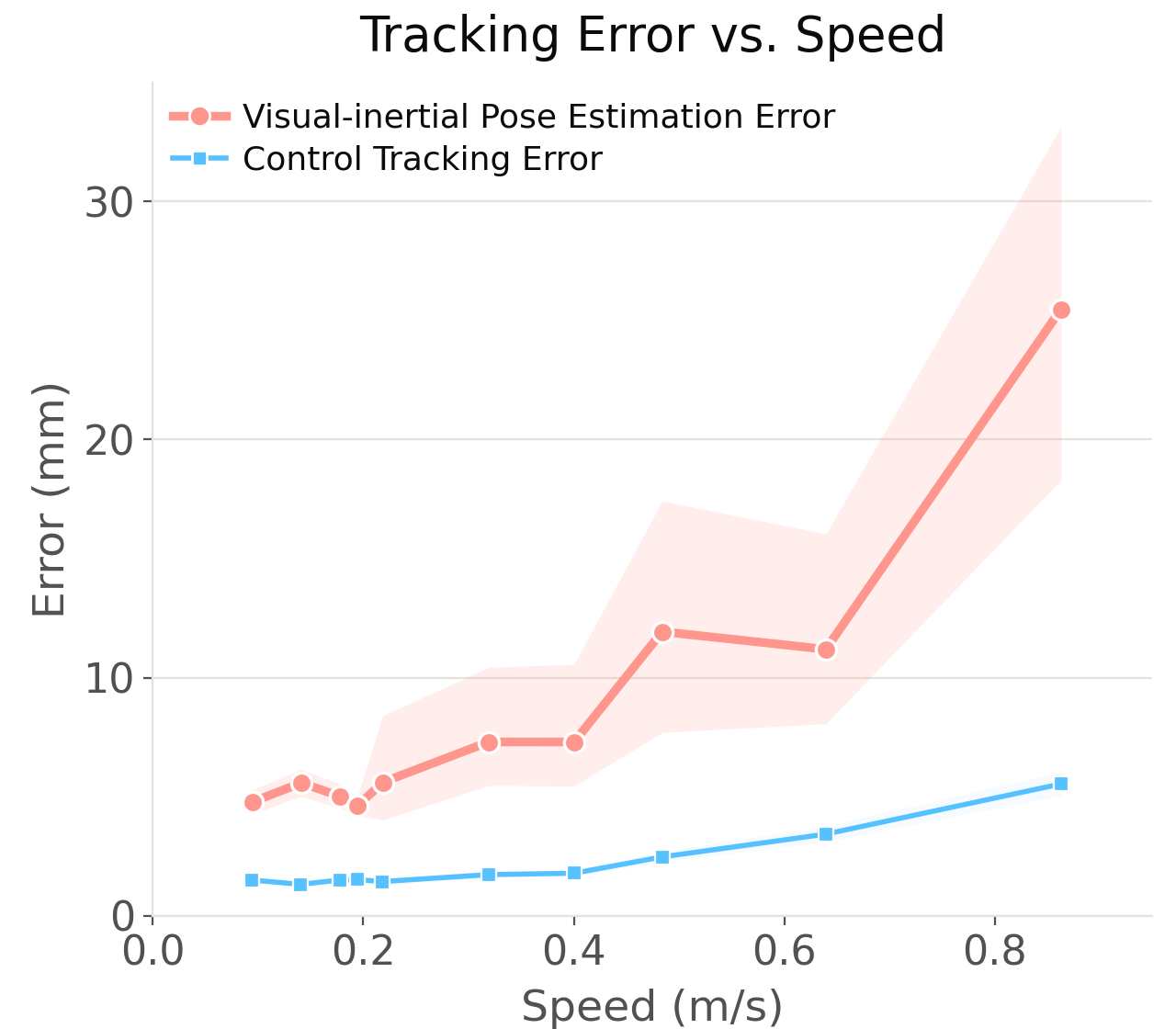}
        \caption{Error vs.\ end-effector speed.}
        \label{fig:vio-tracking}
    \end{subfigure}
    \hspace{0.04\textwidth}
    \begin{subfigure}[b]{0.4\textwidth}
        \centering
        \includegraphics[width=\linewidth]{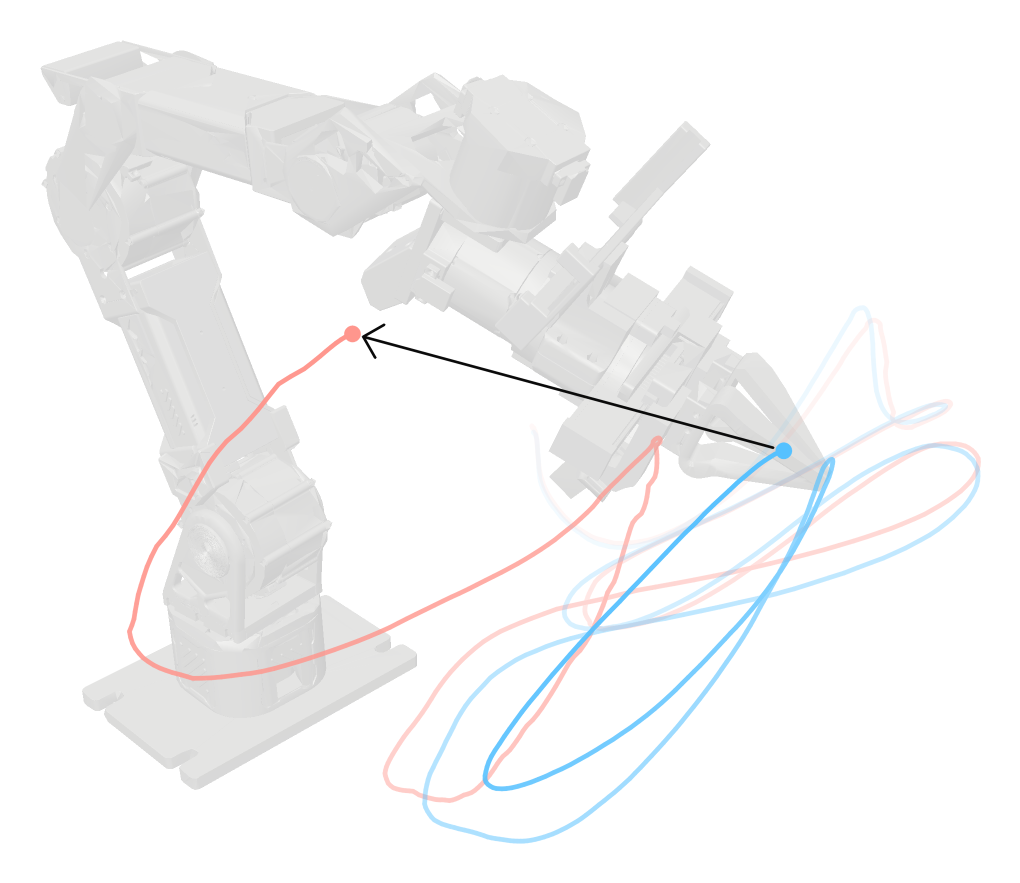}
        \caption{Visual-inertial drift during a fast free-space motion.}
        \label{fig:drift}
    \end{subfigure}
    \caption{\textbf{\textit{Visual-Inertial Pose Estimation vs.\ Control Tracking.}} (a) Average error in 1\,s windows over replayed action trajectories, binned by end-effector speed. (b) End effector trajectory of a recording test from the robot's encoders (blue) and a wrist-mounted visual-inertial tracker (red), a drift of over 10 cm. Per-task policy rollouts and setup details can be found in Appendix~\ref{app:rollout-tracking}, \ref{app:vio-setup}.} 
    \label{fig:vio}
\end{figure*}

Learning from human demonstrations has proven to be a powerful paradigm for teaching robots a wide array of complex skills. We have seen robots learn everything from fine-grained manipulation tasks such as threading a shoelace and lighting a match to mobile manipulation tasks such as sauteing and serving shrimp~\cite{zhao2023learning,fu2024mobile, black2025real}. However, despite these successes, a significant gap remains between robotic and human performance: \textit{speed}. While a person can perform many everyday chores with fluid and rapid motions, policies learned from demonstration typically execute with a slow and deliberate pace that feels distinctly non-human. This disparity is more than just an aesthetic issue; it is a fundamental barrier to deploying robots in practical human-centric environments where efficiency and responsiveness are important. For time-sensitive tasks like cooking and serving food, or collaborative tasks requiring synchronization with humans, the value of a robotic assistant diminishes greatly if it performs these tasks at a fraction of human speed.

The root of this speed gap lies in two challenges. First, there is a lack of accessible systems for collecting demonstration data that capture the dynamics of human-speed motion. Teleoperation interfaces based on puppeteering~\cite{zhao2023learning}, 3D mice~\cite{zhu2020robosuite}, or VR controllers~\cite{zhang2018deep} are not designed for high-speed interaction, due to latency, unintuitive control, and limited sensory feedback for human operators, causing heavy mental load for human operators. This forces human operators to move slowly and carefully, resulting in datasets of sluggish trajectories. The policy, in turn, learns to imitate this slow behavior. Handheld devices  ~\cite{chi2024universal} offer the benefit of more intuitive data collection, and promise scalability by avoiding demonstrations on real robots. However, such devices commonly rely on visual and inertial information for localization, and completely fail in pose estimation for high speed motions (Fig.~\ref{fig:drift}). State of the art visual-inertial tracking systems degrade quickly with speed, seen in Fig.~\ref{fig:vio-tracking}. 

Second, even after assuming near perfect recording, learning a policy that can reliably execute these motions is a significant hurdle. Using recorded positions as actions as in handheld devices \cite{chi2024universal} assumes perfectly reaching every commanded position on time, requiring high torques at high speeds. Discontinuities from action chunking create catastrophic instability. Additionally, position data may not always include all necessary information for forceful manipulation, such as when pressing hard on near stationary objects. Visually, having the operator in view of the camera when directly moving the robot also can confuse the policy.

To address these issues, we introduce \method{}, a practical system for directly learning near human speed bimanual manipulation from human demonstrations. During data collection, the end effectors of lightweight robot arms are attached with two scissor handles for human operators to backdrive the arms under gravity and friction compensation. This method of direct physical guidance provides operators with rich real-time sensory feedback in vision, haptics, and sound, enabling high speed, precise, and contact rich data collection. The arms' motor encoders provide precise position readings even under aggressive motion, avoiding pose estimation issues. To ensure that position trajectories produce effective policies, operators at collection time may directly replay recorded trajectories on the arms between episodes, catching any tracking issues that may arise due to missing action data.

We present a learning pipeline for high-speed manipulation data. Our approach incorporates two key techniques. First, to ensure smooth execution, we introduce test-time smoothing, a post-hoc real-time processing for smooth transitions between action chunks. This method helps mitigate jerkiness at high speed that can arise from action chunking, allowing the robot to perform fast motions without sacrificing stability. Second, to address the visual domain shift from a human operator being present with hands holding the robot end-effectors, we employ a masking technique to align the visual domains between collection and deployment. By removing regions with human hands appearing in cameras via an analytical model, we align the visual input distributions. We find that while performance remains high without visual alignment, masking further improves performance.


The synergy between our intuitive hardware for near-human-speed data collection and our robust learning pipeline allows \method{} to learn fast, complex tasks from a small number of demonstrations. With only 50 demonstrations per task, our system learns to autonomously perform fine-grained tasks such as tossing a ball from one arm and catching it with the other, folding chino shorts, bussing a table, inserting a battery into a remote, and unstacking plastic cups, achieving over 80\% success rates. Critically, these tasks are completed at speeds that are close to average human speeds, reducing the execution time by over 50\% compared to that of teleoperation systems, and represent a significant step towards creating robots that can operate efficiently in the real world.


\section{Related Work}
\noindent \textbf{Learning from Demonstrations.}
Learning from demonstrations, or imitation learning, is a cornerstone of modern robotics, enabling robots to acquire complex skills by mimicking expert demonstrations~\cite{billard2008robot}. The efficacy of this paradigm is heavily dependent on the interface used for data collection. A wide spectrum of interfaces has been explored, ranging from processing human videos to various forms of interactive teleoperation. While human videos offer immense scale and diversity, they introduce significant embodiment and observation gaps that complicate direct policy learning~\cite{kareer2025egomimic,lepert2025phantom,zhu2025emma}. Teleop methods aim to reduce this gap. Common low-cost interfaces include keyboards~\cite{xiong2024adaptive}, 3D space mice~\cite{zhu2023viola}, game controllers~\cite{ross2013learning} and phones~\cite{mandlekar2018roboturk}, though they can be unintuitive for controlling 6-DoF motion. More immersive interfaces, such as Virtual Reality controllers, provide more direct 6-DoF end-effector control but can suffer from latency, a lack of haptic feedback, and a lack of feedback on kinematic constraints~\cite{zhang2018deep,cheng2024tv,park2024dexhub,seo2023deep}. Other approaches shadow human motion in real time or offline using real-time pose estimation~\cite{handa2020dexpilot,sivakumar2022robotic,fu2024humanplus,he2024learning}, exoskeletons~\cite{fang2024airexo,purushottam2023dynamic} and vision-based tracking~\cite{chi2024universal,wang2024dexcap,etukuru2025robot}, capturing natural kinematics but still facing the human-to-robot correspondence problem. To achieve a more direct mapping, 
leader-follower puppeteering systems provide secondary leader devices that the operator manipulates to control the follower robots providing intuitive joint-space control~\cite{jenness1975master,zhao2023learning,fu2024mobile,wu2024gello,buamanee2024bi,ito2022efficient,jiang2025behavior}. In contrast to these methods, \method{} utilizes kinesthetic teaching where human operators directly hold and guide the robot motion, eliminating the cost and complexity of a duplicate leader setup for puppeteering and bypassing the latency or embodiment gap inherent in other interfaces.

\noindent \textbf{Kinesthetic Teaching.}
Kinesthetic teaching, where an operator physically guides a robot's end-effector through a task, is one of the most direct and intuitive methods 
\noindent for programming robot behaviors~\cite{billard2008robot}. By moving the robot by hand, the demonstrator can implicitly convey not only the desired trajectory but also information about contact points and forces. This method requires the robot to be either passively compliant or actively backdrivable, often necessitating gravity compensation to make the robot feel ``weightless'' and reduce operator fatigue. Early work demonstrated the effectiveness of this approach for learning simple trajectories and dynamic movement primitives~\cite{finn2016guided,xie2019improvisation,akgun2012trajectories,kormushev2011upper,kormushev2011imitation,weiss2009teaching,ben2009kinesthetic}. However, a primary challenge has been the difficulty of recording fast, dynamic motions due to robot inertia and motor friction, which can make the robot feel sluggish and heavy to the operator. Furthermore, the physical presence of the human operator in the workspace during demonstration can create a visual domain shift, as the policy is later deployed in a scene without the operator's hands and arms. \method{} addresses the challenge of collecting high-speed data by using a lightweight backdrivable bimanual platform with low-gear-ratio actuators with real-time gravity compensation along with an ergonomic scissor handle design, significantly reducing operator effort. Optional motor friction compensation and gripper gear upsizing further reduce operator fatigue and improve trajectory quality. Together with our masking and real-time smoothing, \method{} can capture and learn end-to-end policies for precise and fast skills from fluid, near-human-speed motions that are difficult to obtain and execute with kinesthetic teaching systems.

\noindent \textbf{High-Speed Manipulation.}
Achieving human-level speed in manipulation is a long-standing goal in robotics, crucial for applications where efficiency is paramount. Traditional approaches to high-speed robotics often rely on precise analytical models of robot and object dynamics, using techniques like trajectory optimization to generate feasible, time-optimal paths. These methods have enabled impressive feats, such as industrial robots performing rapid pick-and-place operations or research platforms batting~\cite{senoo2006ball}, playing table tennis~\cite{dambrosio2025achieving,buchler2022learning,mulling2013learning,zhu2018towards,liu2013racket}, moving objects and avoiding collisions at speed~\cite{ichnowski2022gomp}, tossing~\cite{zeng2020tossingbot}, dynamically manipulating deformable objects~\cite{zhang2021robots,chi2024iterative,ha2022flingbot}, and catching objects in mid-air~\cite{kim2014catching}. However, these approaches typically require extensive, task-specific engineering and can be brittle to unmodeled contact dynamics or perturbations. Learning-based methods offer a promising alternative by acquiring 
dynamic skills directly from data. However, a primary bottleneck has been the lack of systems capable of collecting demonstration data that captures near-human-speed motion~\cite{masuya2025variable,guo2025demospeedup}. 

Furthermore, even with a suitable dataset, learning a policy to robustly execute fast motions is a significant hurdle~\cite{chi2024universal,arachchige2025sail,yuan2025speedtuning}. Fast movements are less forgiving, where latency discrepancies can cause out-of-sync actions, and discontinuous commands from a policy can lead to instability. 

Orthogonal to collecting high speed human data, reinforcement learning may also be used to collect high speed data while preserving effectiveness. Such approaches typically explicitly model speed as an objective \cite{arachchige2025sail, guo2025demospeedup, kim2026espada}. Other speedup approaches also include using different action representations that provide analytical representations of trajectories for speedup ~\cite{han2026b}. \method{} is complementary to these approaches, by allowing operators to directly train high speed policies without refinement.

\section{Hardware Setup \& Data Collection}
We set up our hardware with the consideration of using kinesthetic teaching for fast and precise data collection. We prioritize comfort for human operators, pursuing maximal achievable speed using our hardware setup. 

\begin{figure}[t]
    \centering
    \includegraphics[width=\linewidth]{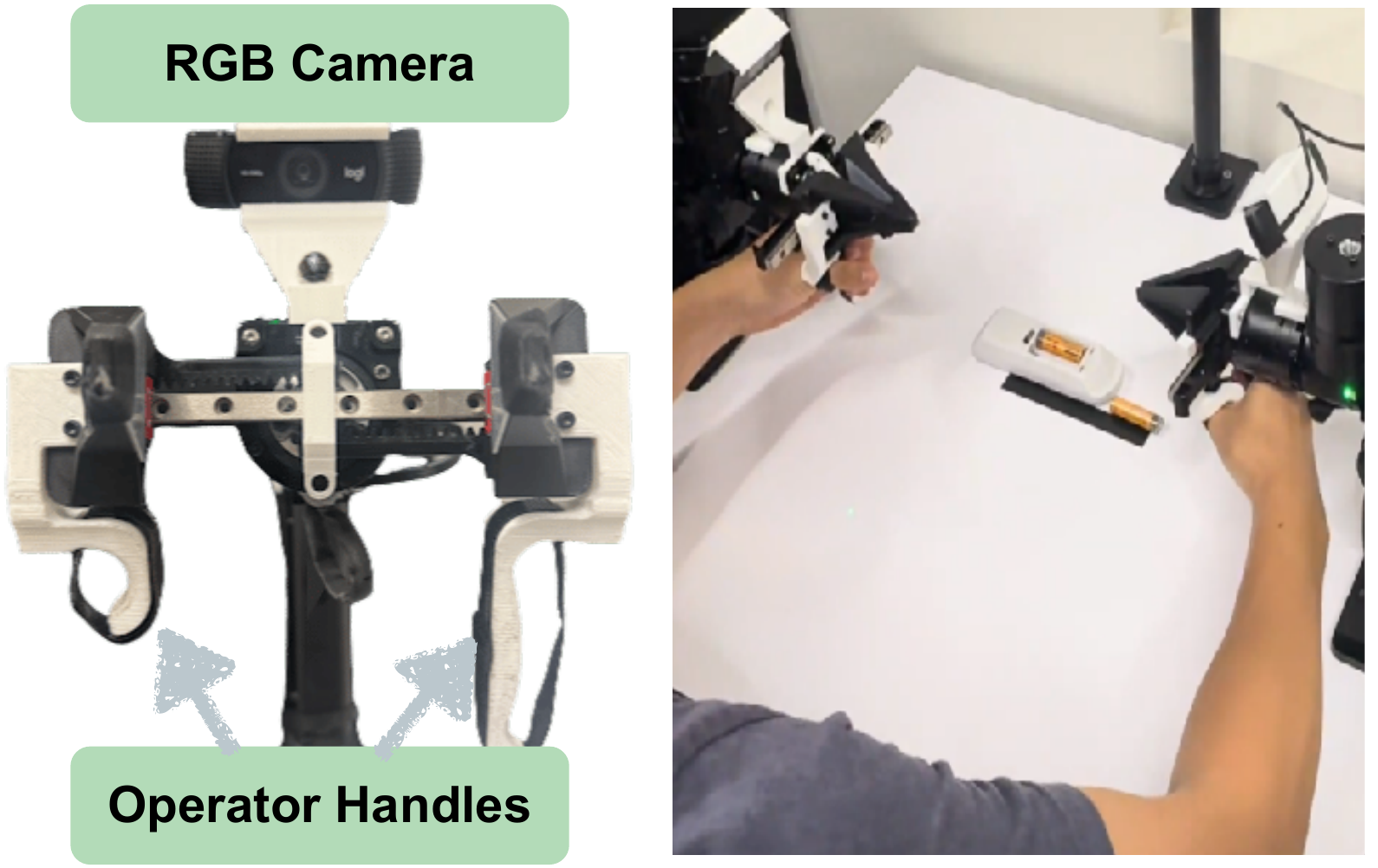}
    \caption{\textbf{\textit{Hardware Setup.}} Our printed handles allow for easy operation and data collection.}
    \label{fig:hardware-setup}
\end{figure}

Shown in Fig.~\ref{fig:hardware-setup}, \method{} consists of a bimanual setup of two different types of 6-DoF robot arms, HESSIAN and ARX, each with one parallel jaw gripper. We leverage off-the-shelf robot arms using planetary actuators, and human operators can back-drive the robot arms during kinesthetic teaching with relatively low internal friction in the joints, compared to others like harmonic actuators. We only use the HESSIAN setup for the highly dynamic Toss \& Catch task since this task requires high joint velocity limits for high-speed trajectory following and very low joint friction for high-speed data collection. 

We implement real-time gravity compensation based on the joint positions and masses of each robot link through torque control at 100Hz~\cite{lynch2017modern}. We retrofit the end effector of a robot arm with scissor handles for human operators to hold and backdrive the robots arms for completing manipulation tasks. One operator handle is attached on each jaw of the gripper. The smaller handle is for the thumb, and the other handle is for the other four fingers. Fingers are held in place with an adjustable strap, providing human operators with firm and intuitive grasps of the end effectors of the robot arms. The long finger handle allows operators to pull downwards and sideways, while hand contact with the bottom of the arm assists with upwards acceleration. The whole setup consists of three RGB cameras capturing images at 50Hz, with one wrist camera mounted on each arm facing towards the region between the gripper jaws, and one top-down camera capturing the workspace from above. 

During data collection, we record three 360 x 640 RGB images $I_1, I_2, I_3$ from the three cameras, joint positions $q$, and joint velocities $\dot{q}$ at 50Hz. Through kinesthetic teaching where the human operators directly hold and guide the robots to do different tasks, human operators obtain various real-time sensory feedbacks, including 3D vision, haptics through the handles, and contact sound without any delays that exist in common teleoperation platforms.

\section{Data Processing, Policy Learning \& Deployment}
Given a dataset of kinesthetic teaching data for fast and precise manipulation, directly applying imitation learning algorithms fail for two reasons. First, a human operator is present with hands holding the robot end-effectors during data collection, while the robots are alone during the autonomous execution phase after deployment. This creates a visual distribution shift between data collection and deployment. Second, fast movements are less forgiving, as small errors in execution can compound into unrecoverable failures and non-smooth actions can cause instability. The high-speed nature of the manipulation data collected presents a hurdle for learning a reliable policy. 

\begin{figure}[t]
    \centering
    \includegraphics[width=\linewidth]{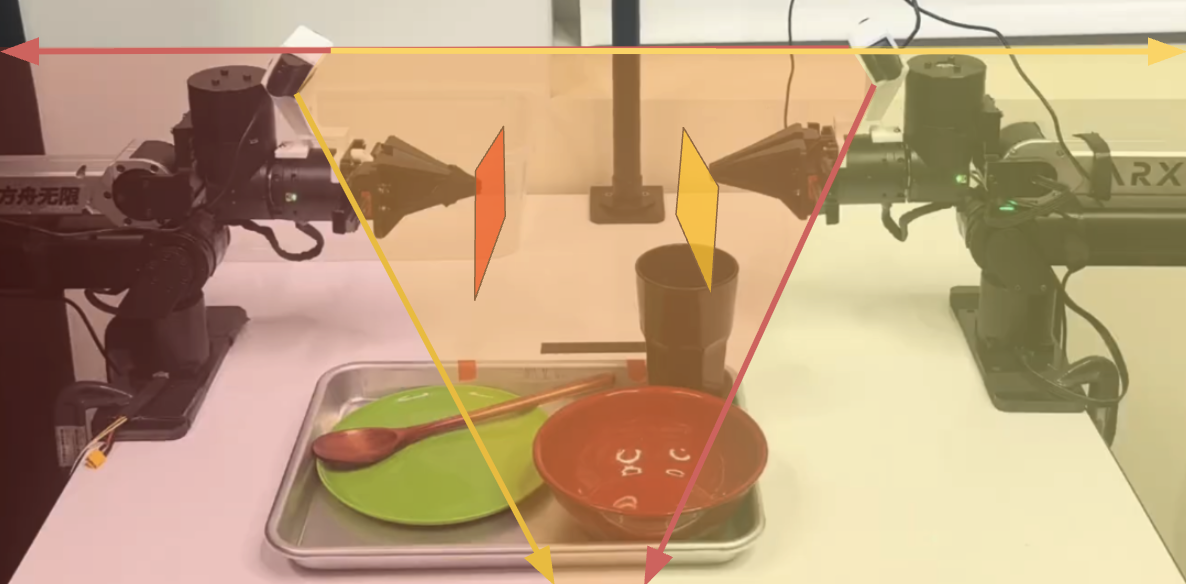}
    \caption{\textbf{\textit{Masking.}} Our masking method adjusts in real time based on the position of the end effectors of both arms.}
    \label{fig:mask}
\end{figure}

\begin{figure}[t]
    \centering
    \includegraphics[width=\linewidth]{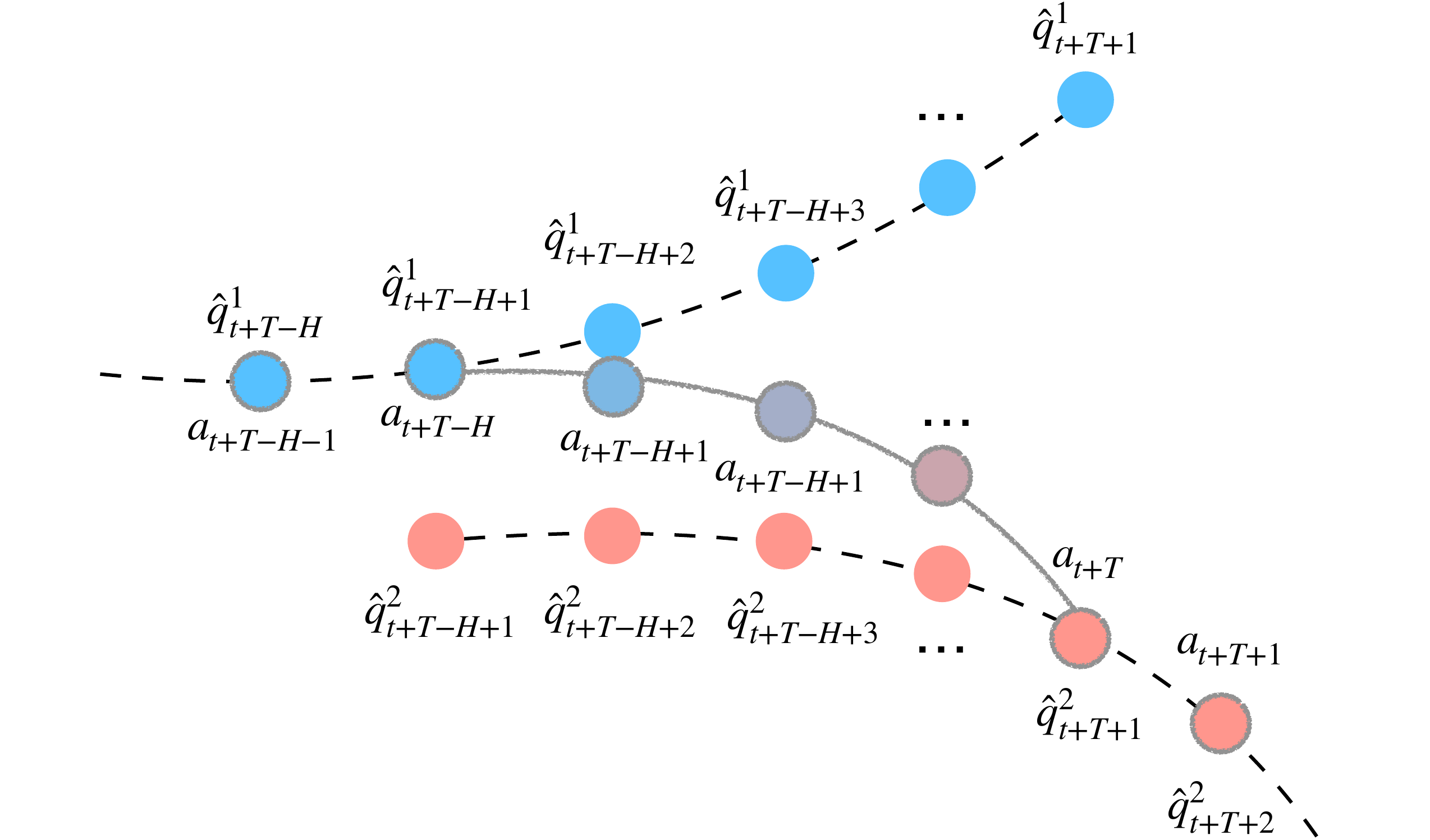}
    \caption{\textbf{\textit{Test-Time Smoothing.}} Given a previous chunk of target joint positions $\hat{q}^{1}$ and a later chunk $\hat{q}^{2}$, we apply smoothing to obtain the final commanded actions $a$.}
    \label{fig:smoothing}
\end{figure}

To address the distribution shift between data collection and deployment, we employ a masking technique to align the visual domains for training and testing, illustrated in Fig.~\ref{fig:mask}. By removing regions with human hands and arms appearing in cameras using an analytical model, we align the visual input distributions, preventing the policy from being distracted by the presence or absence of the human operator. And to ensure smooth execution of fast manipulation, we introduce test-time smoothing, a post-hoc real-time processing for smooth transitions between action chunks. Illustrated in Fig.~\ref{fig:smoothing}, this method helps mitigate the jerkiness at high speed that can arise from action chunking~\cite{black2025real}, allowing the robot to perform fast motions without sacrificing stability. 

\noindent \textbf{Masking in Image Space.}
During training, we pre-process the collected data, aiming to mask out the appearances of hands and arms of the human operators. Given that the joint positions of both arms and robot description files are known, we can compute the poses of end-effectors of both robot arms in their respective base frames using forward kinematics ($\text{FK}$). We also know the transformation between the base frames of both arms through camera calibration. Thus, we can transform the end-effector pose of one arm into the base frame of the other arm. Since the relative pose of the handles for the human hands is fixed with respect to the end-effector pose, we deterministically calculate the center pose of the left hand in the right camera, and similarly for the right hand in the left camera, and mask out a small square region perpendicular to the end-effector pointing direction near the handles by setting the image RGB values for that region to be zero. For the top camera, we similarly mask out two rectangular regions for both hands and arms based on the end-effector poses. We provide an illustration for masking in Fig.~\ref{fig:mask}. During deployment, we perform masking in real time for every frame. Formally, at each time step $t$, we perform the masking operation $M$ and get three processed images $I'_{1,t}, I'_{2,t}, I'_{3,t}$, where
\[
I'_{1,t}, I'_{2,t}, I'_{3,t} = M(I_{1,t}, I_{2,t}, I_{3,t}, q_t, \text{FK}).
\]

\noindent \textbf{Policy Learning. }
Unlike teleoperation where the action commands for the robots are explicitly sent through devices like leader arms or VR controllers, we only capture the visual and proprioceptive readings from the robots. We aim to train policies that can predict future joint positions which the robots should achieve given current images and proprioception. We follow the recent progress in imitation learning by using action chunking~\cite{zhao2023learning}, which mitigates compounding error problems that is common in behavior cloning methods and improves overall performance. Given a policy $\pi$ and an action chunk size of $T$, we train the policy by maximizing the probability
\[
\pi([q_{t+1}, ..., q_{t+T}] | I'_{1,t}, I'_{2,t}, I'_{3,t}, q_t),
\]
where the predicted future joint positions will serve as the target joint positions for a low-level PD controller to execute. We additionally tune an action lookahead where we skip the first $n$ actions at every chunk boundary, resulting from action delay at execution. In this paper, we use ACT~\cite{zhao2023learning} as the policy representation, but we expect other policy representations for imitation learning with action chunks would also work. 

\noindent \textbf{Test-Time Smoothing. }
When moving at increased speeds, we observe that discontinuities between action chunks~\cite{black2025real} become larger and more visible, affecting policy performance. We introduce a simple post-hoc test-time smoothing technique. Illustrated in Fig.~\ref{fig:smoothing}, we tackle the discontinuity between two adjacent chunks by blending 
the last few actions of a chunk with the first few actions of the later chunk where these two action chunks have a small overlap in corresponding timesteps. For a smoothing window size of $H$ smaller than the action chunk size $T$, an action chunk $[\hat{q}^1_{t+1}, ..., \hat{q}^1_{t+T}]$ is first predicted at timestep $t$, and another one $[\hat{q}^2_{t+T-H+1}, ..., \hat{q}^2_{t+2T-H}]$ is later predicted at timestep $t+T-H$. At timestep $t'$, the action command $a_{t'}$ is the raw predicted target joint position $\hat{q}_{t'+1}$ at regions where there is no chunk overlap. Within the overlapping regions, we weigh the two corresponding target joint positions by using a sigmoid coefficient $w(t') = \text{sigmoid}\big((t' - (t+T-H) - H/2) / \sigma\big)$ where $\sigma$ is a time scale, and the final blended action command is 
\[
a_{t'} = (1 - w(t')) * \hat{q}^1_{t'+1} + w(t') * \hat{q}^2_{t'+1}.
\]
By adjusting the smoothing horizon and time scale of the sigmoid coefficient, we can adjust how sharply and reactively the robot will change to follow a later action chunk from a previous obsolete action chunk.


\section{Experiments}
\begin{table*}[t]
\setlength{\tabcolsep}{2pt}
\renewcommand\arraystretch{1.0}

\resizebox{\linewidth}{!}{
\begin{tabular}{@{}l
*{4}{c} *{2}{c}  
*{6}{c}          
*{4}{c}          
*{4}{c}          
@{}}
\toprule
\multicolumn{19}{c}{\textbf{Masking}}\\
 & \multicolumn{4}{c}{\textbf{Unstack Cups}} & \multicolumn{6}{c}{\textbf{Battery Insertion}} & \multicolumn{4}{c}{\textbf{Bus Table}} & \multicolumn{4}{c}{\textbf{Fold Shorts}} \\
Method
& \makecell{Grab } & \makecell{Pull} & \makecell{Whole} & \makecell{Time(s)}
& \makecell{Pick} & \makecell{Place} & \makecell{Prepare} & \makecell{Push} & \makecell{Whole} & \makecell{Time(s)}
& \makecell{C\&S} & \makecell{B\&P} & \makecell{Whole} & \makecell{Time(s)}
& \makecell{Up} & \makecell{Right} & \makecell{Whole} & \makecell{Time(s)} \\
\midrule
No Mask
& 90 & \textbf{90} & \textbf{90} & 4.12
& 95 & 95 & 85 & 75 & 75 & 6.44
& \textbf{100} & \textbf{100} & \textbf{100} & 6.97
& \textbf{100} & 90 & 90 & 6.19 \\
Inpainting
& \textbf{100} & 85 & 85 & \textbf{3.92}
& 95 & 95 & 70 & 65 & 65 & \textbf{5.65}
& \textbf{100} & \textbf{100} & \textbf{100} & \textbf{6.57}
& \textbf{100} & 80 & 80 & 6.20 \\
\textbf{Ours}
& \textbf{100} & \textbf{90} & \textbf{90} & 3.97
& \textbf{100} & \textbf{100} & \textbf{90} & \textbf{85} & \textbf{85} & 5.70
& \textbf{100} & \textbf{100} & \textbf{100} & 6.81
& \textbf{100} & \textbf{100} & \textbf{100} & \textbf{5.94} \\
\bottomrule
\end{tabular}
}
\caption{\textbf{\textit{Our masking improves performance. }}Our masking strategy almost uniformly beats non-masked and diffusion inpainted in policies, enabling tasks such as battery insertion.}
\label{tab:masking}
\end{table*}
In our experiments, we aim to determine the extent to which our masking strategy as well as our smoothing method benefit the overall success rate of policies. Additionally, we study the effects of several comparative approaches in these two domains. Through user trials, we verify the intuitive understanding that our directly operated data collection platform enables faster training of human operators. All main tasks are shown in Fig.~\ref{fig:teaser}.

\subsection{Tasks}
In \textbf{\textit{Unstack Cups}}, the left arm reaches out and grasps the bottom of two stacked red solo cups. Then, the right arm subsequently grasps the lip of the top cup without contacting the bottom, and pulls. The grippers must carefully balance their grasps; if the left grips too hard, the right can't tilt and grab. If the right over-grips, it pulls the entire stack. The cups' initial starting positions are randomized along a fifteen centimeter piece of tape. This task is collected with a timer of four seconds. 

\textbf{\textit{Battery Insertion}}. During execution, one arm picks up a battery and places it at an oblique angle into a remote control, taking care not to touch the spring. It then pushes one end of the battery into the spring. The other gripper then raises and forcefully pushes down on the battery, slotting it into place. The battery and the remote are randomized along the tape. This task is collected in under seven seconds. 

\textbf{\textit{Bus Table}} involves removing common dining ware from a pan and placing them into a tray. An arm first grasps a nearby tray and brings it closer to the pan. Then, both arms engage, alternating picking up and placing four items from the pan into the tray: a cup, a bowl, a spoon, and a plate. The position of the pan is randomized along the tape. This task is collected with a timer of six seconds.

\textbf{\textit{Fold Shorts}} involves folding a pair of shorts placed flat. The arms must reach and grasp the edge of the bottom trunk of the shorts and toss the trunk over from bottom to top flush against the top cover. Then, the left gripper must grasp the edge of the shorts along the waist, and toss it from left to right while the right arm creates a crease. The shorts are randomly rotated by up to thirty degrees in their initial position, and this task is collected against a timer of five seconds.

\textbf{\textit{Toss \& Catch}}. The left arm first approaches and picks up a small soft ball on the table. Then it swings forward and up dynamically in high speed while releasing the ball. The right arm approaches the tossed ball-in-air swiftly with closing grippers to catch the ball. The task is only two seconds long.

\subsection{Comparisons}
We compare our method to \textit{Inpainting}, \textit{No Mask}, \textit{Temporal Ensembling} (TE), \textit{No Smooth}, \textit{Puppeteering}, \textit{Puppet Double Freq}, and \textit{Puppet Skip Actions}.

\textbf{Inpainting~\cite{xu2025dexumi}.} A policy trained on diffusion inpainted video, where the operator hands are segmented and replaced with the background by a pretrained segmentation~\cite{ravi2024sam2} and pretrained diffusion network~\cite{zhou2023propainter}. 

\textbf{No Mask.} A policy trained with the raw video.

\textbf{Temporal Ensembling~\cite{zhao2023learning}}. A post-process smoothing method where an action chunk is predicted every timestep, and exponentially weighted and averaged with past action chunks.

\textbf{No Smooth.} No smoothing method applied.

\textbf{Puppeteering~\cite{zhao2023learning}.} The operator collects and deploys policies on a teleoperated setup.

\textbf{Puppet Double Freq.} On the puppeteering setup, the policy is deployed with double the controller frequency.

\textbf{Puppet Skip Actions.} On the puppeteering setup, the policy acts on every other predicted action per chunk.

\subsection{Masking Aligns Data Collection and Deployment}

We evaluate the efficacy of masking, which aligns the visual distribution of the policy between collection and deployment time. Table \ref{tab:masking} details the results of our experimental trials comparing policies trained with masking, no masking, and on diffusion inpainted video. We note an increase in performance on most tasks with masking compared to simply training with hands included and inpainted video.

\begin{figure}[t]
    \centering
    \includegraphics[width=\linewidth]{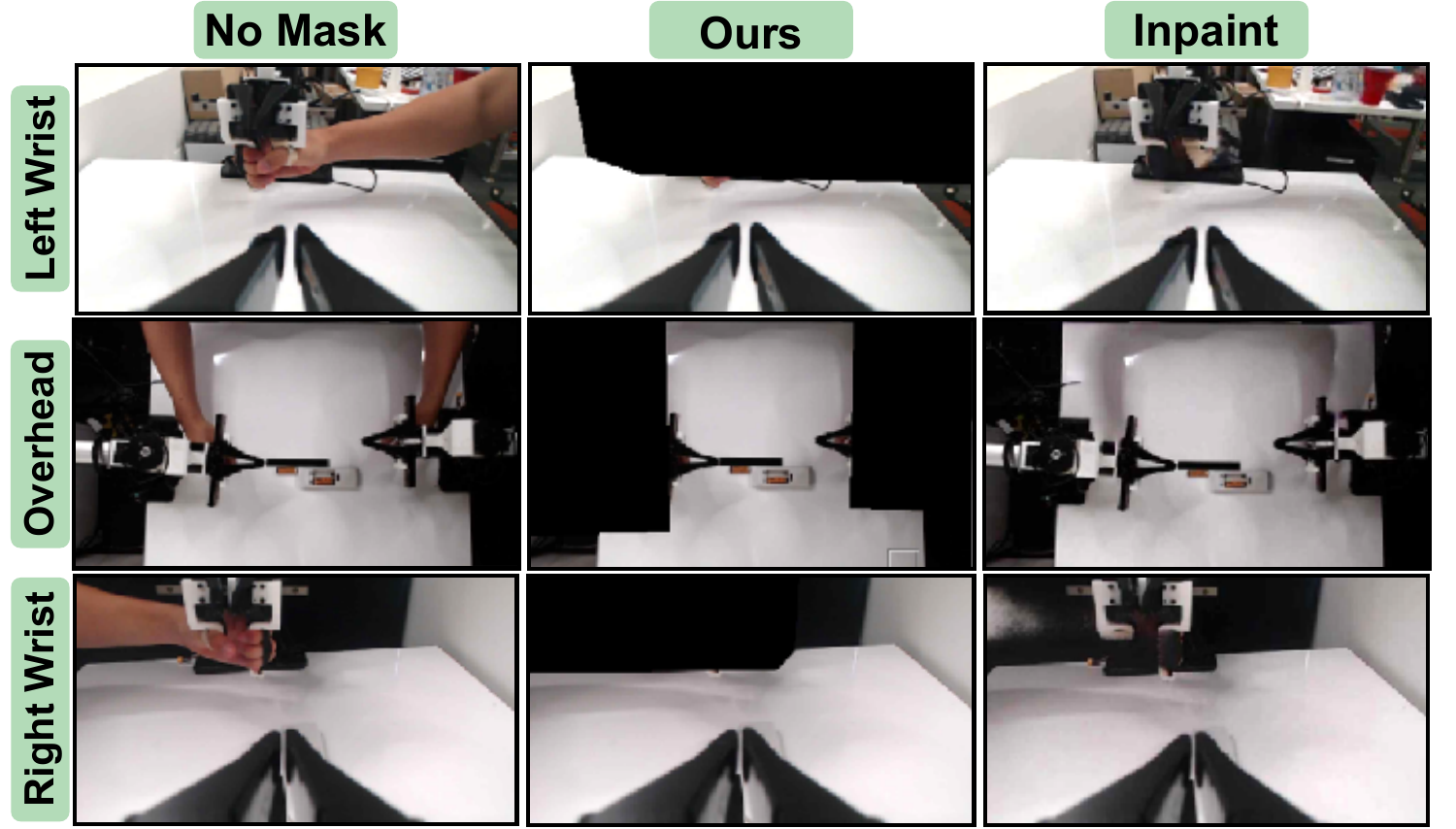}
    \caption{\textbf{Masking Comparisons.} We show the same initial starting frame of \textbf{\textit{Battery Insertion}}. Our masking method completely aligns the visual distributions of training and test times. Inpainting via diffusion creates unwanted artifacts.}
    \label{fig:alignment}
\end{figure}

The policies trained on inpainted video suffer significant performance losses compared to the base video. We hypothesize that this is due to two factors. First, as can be seen in Fig.~\ref{fig:alignment}, the shadows of the arms are still visible against the reflective surface of the table. Another factor is the noisy diffusing of frames due to lack of context. These visual artifacts not visible at test time may confuse the policy. Given a more controlled environment, this method's performance may improve.

\subsection{Test-Time Smoothing Improves Performance}
\begin{table*}[t]
\resizebox{\linewidth}{!}{
\setlength\tabcolsep{2.5pt}
\begin{tabular}{@{}l
*{4}{c} *{6}{c} *{4}{c} *{4}{c}
@{}}
\toprule
\multicolumn{19}{c}{\textbf{Smoothing}}\\
 & \multicolumn{4}{c}{\textbf{Unstack Cups}} & \multicolumn{6}{c}{\textbf{Battery Insertion}} & \multicolumn{4}{c}{\textbf{Bus Table}} & \multicolumn{4}{c}{\textbf{Fold Shorts}} \\
Method
& \makecell{Grab} & \makecell{Pull} & \makecell{Whole} & \makecell{Time(s)}
& \makecell{Pick} & \makecell{Place} & \makecell{Prepare} & \makecell{Push} & \makecell{Whole} & \makecell{Time(s)}
& \makecell{C\&S} & \makecell{B\&P} & \makecell{Whole} & \makecell{Time(s)}
& \makecell{Up} & \makecell{Right} & \makecell{Whole} & \makecell{Time(s)} \\
\midrule
No Smooth
& \textbf{100} & 65 & 65 & 4.23
& \textbf{100} & 85 & 75 & 75 & 75 & \textbf{5.29}
& \textbf{100} & 80 & 80 & \textbf{6.70}
& 95 & 80 & 80 & 6.16 \\
TE
& 95 & 35 & 35 & 7.75
& \textbf{100} & \textbf{100} & 85 & 55 & 55 & 10.18
& \textbf{100} & \textbf{100} & \textbf{100} & 12.23
& \textbf{100} & 90 & 90 & 11.60 \\
\textbf{Ours}
& \textbf{100} & \textbf{90} & \textbf{90} & \textbf{3.97}
& \textbf{100} & \textbf{100} & \textbf{90} & \textbf{85} & \textbf{85} & 5.70
& \textbf{100} & \textbf{100} & \textbf{100} & 6.81
& \textbf{100} & \textbf{100} & \textbf{100} & \textbf{5.94} \\
\bottomrule
\end{tabular}
}
\caption{\textbf{\textit{Smoothing improves performance.}} Test time smoothing outperforms or matches other methods in success rate in every category.}
\label{tab:smoothing}
\end{table*}

In Table~\ref{tab:smoothing}, we see performance comparisons between non-smoothed, test-time smoothed, and temporal ensembling-smoothed trajectories. In all tasks, test-time smoothing achieves the best performance, and by a wide margin in tasks \textbf{\textit{Unstack Cups}} and \textbf{\textit{Battery Insertion}}. The resulting instabilities in the high speed non-smoothed policy rollouts result in imprecise grasps and pushes, and while the temporal-ensembling smoothed policies have mixed results, their execution speeds drop in half dramatically from collection time. This is due to the increased processing time necessary to compute inference every timestep.

A failure mode of temporal ensembling is its tendency to choose an averaged path when forced to change course \cite{black2025real}. We hypothesize that, at higher speeds, the tracking of robot controllers often become less reliable which forces policies to react and frequently change course to adjust for error. Thus, even if temporal ensembling were run on faster hardware and had no inference delay, it would perform even more poorly at speed due to this phenomenon. This also partially explains the presence of discontinuities in the first place, even in relatively uni-modal data with little strategy-switching.

\subsection{Speed Comparisons}

In this experiment, we investigate the effectiveness of training a policy on slow data, collected with puppeteering on an ALOHA~2 (setup in Appendix~\ref{app:aloha2}). We artificially increase the policy execution speed in two different ways across three different tasks: \textbf{\textit{Bus Table}}, \textbf{\textit{Unstack Cups}}, and \textbf{\textit{Battery Insertion}}. We are unable to complete the \textbf{\textit{Toss \& Catch}} task with puppeteering due to the difficulty of catching a dynamic object through teleoperation. The results are visible in Table \ref{tab:puppeteering}, where we see sharp drops in performance after speeding up data collected through puppeteering to match the speed of ours. Refer to Appendix~\ref{app:chunk-jump} for chunk boundary jump details.

\begin{table*}[t]
\resizebox{\linewidth}{!}{
\setlength\tabcolsep{2.0pt}
\begin{tabular}{@{}l
*{4}{c} *{6}{c} *{4}{c} *{4}{c}
@{}}
\toprule
\multicolumn{19}{c}{\textbf{Speed Comparisons}}\\
 & \multicolumn{4}{c}{\textbf{Unstack Cups}} & \multicolumn{6}{c}{\textbf{Battery Insertion}} & \multicolumn{4}{c}{\textbf{Bus Table}} & \multicolumn{4}{c}{\textbf{Toss \& Catch}} \\
Method
& \makecell{Grab} & \makecell{Pull} & \makecell{Whole} & \makecell{Time(s)}
& \makecell{Pick} & \makecell{Place} & \makecell{Prep} & \makecell{Push} & \makecell{Whole} & \makecell{Time(s)}
& \makecell{C\&S} & \makecell{B\&P} & \makecell{Whole} & \makecell{Time(s)}
& \makecell{Pick \& Toss} & \makecell{Catch} & \makecell{Whole} & \makecell{Time(s)} \\
\midrule
Puppeteering
& 85 & 75 & 75 & 8.84
& \textbf{100} & 90 & 60 & 55 & 55 & 17.87
& \textbf{100} & 90 & 90 & 16.05
& -- & -- & -- & -- \\
Puppet Double Freq
& 80 & 65 & 65 & 5.47
& 80 & 65 & 35 & 15 & 15 & 12.90
& 95 & 85 & 85 & 9.30
& -- & -- & -- & -- \\
Puppet Skip Actions
& 85 & 50 & 50 & 5.58
& 95 & 95 & 55 & 30 & 30 & 10.30
& 95 & 85 & 85 & 9.29
& -- & -- & -- & -- \\
\textbf{Ours}
& \textbf{100} & \textbf{90} & \textbf{90} & \textbf{3.97}
& \textbf{100} & \textbf{100} & \textbf{90} & \textbf{85} & \textbf{85} & \textbf{5.70}
& \textbf{100} & \textbf{100} & \textbf{100} & \textbf{6.81}
& \textbf{100} & \textbf{80} & \textbf{80} & \textbf{1.85} \\
\bottomrule
\end{tabular}
}
\caption{\textbf{\textit{Speed Comparisons. }}We note a sharp drop in performance on precise tasks when applying either speed-up approach.}
\label{tab:puppeteering}
\end{table*}

\textbf{Puppet Double Freq} simply doubles the controller frequency at execution time. This results in policies that appear out of control, with large jumps at chunk boundaries. With actions that jump larger distances, such as in \textbf{\textit{Bus Table}}, this effect becomes more noticeable. Despite still moving slower than our method, double-timed policies from puppeteering have significantly larger discontinuities than in our method. We hypothesize that these discontinuities lead to deviations between collection and test time, further worsening performance.

\textbf{Puppet Skip Actions} is running the controller frequency at the same rate, but skipping every other action. This method introduces less discontinuity, but creates new failure modes such as oscillating above the battery instead of pushing it in, and missing grasps. This is due to the policy oftentimes skipping critical steps, which also reduces overall performance.

\subsection{User Study}

We conduct a user study with seven volunteers with little to no teleoperation experience, having them operate our setup and puppeteering. For the study, we choose \textbf{\textit{Battery Insertion}}, a task requiring precision. 

In the study, in both setups, users spend three minutes familiarizing themselves with the objects in the scene and the robotic arms. Next, they are shown an expert demonstration and given basic tips on operation procedures. Finally, they are timed in ten trials, attempting to insert the battery into the remote as quickly as possible.

\begin{figure}[t]
    \centering
    \includegraphics[width=\linewidth]{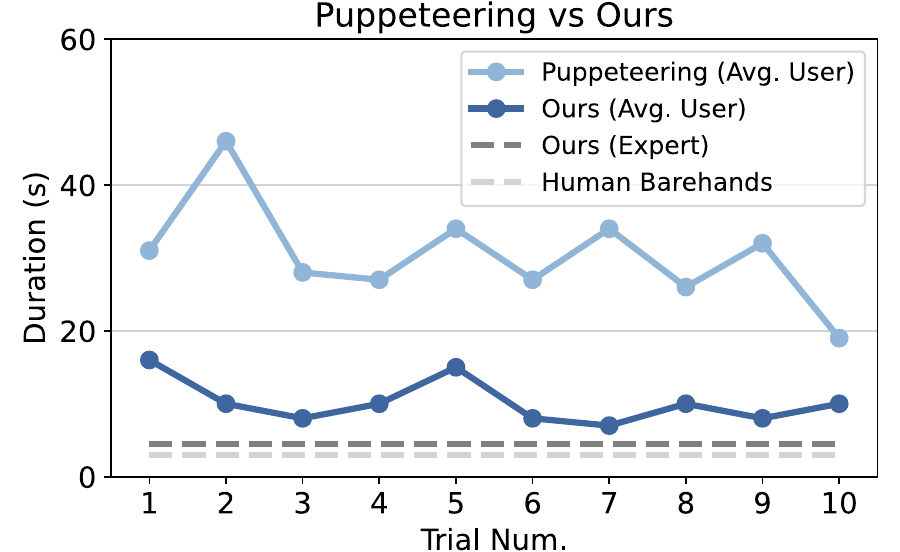}
    \caption{\textbf{\textit{User Study.}} Results on \textbf{\textit{Battery Insertion}} for timing operators on both Ours and Puppeteering demonstrate a 50\%+ gap in duration for both experienced and new operators.}
    \label{fig:user}
\end{figure}

We observe in Fig.~\ref{fig:user} an ability for users to quickly learn how to operate with our hardware and at speeds up to three times as fast as puppeteering. This primarily results from the operator sitting closer to the environment, giving more direct visual feedback, and haptic sensations from operating the actual robot arms. This results in more natural data collection, even for beginners.

\section{Conclusion}
In this paper, we present \method{}, a full-stack system for learning fast and precise manipulation. Our system presents a hardware setup for collecting high-speed demonstration, and a learning pipeline for smooth autonomous execution of fast and precise movements. Currently, the masking strategy that we use in the paper over-conservatively masks regions near human hands and arms, erasing potentially useful visual cues near human hands in the image space. Additionally, the whole system incorporates vision and proprioception as only two modalities, preventing the system from learning dexterous manipulation requiring contact-rich perception and control. In the future we would like to research adaptive masking strategies and including tactile sensing for better performance.

\section*{Acknowledgments}
We'd like to thank Ajay Sridhar, Yanjiang Guo, Brian Kim, Lars Ankile, Marcel Torne, Aneesh Muppidi, Ke Wang, Yuejiang Liu, Emily Jin, and Moo Jin Kim for helpful insights, discussions, and help filming. This work was supported by the Robotics and AI Institute. Zipeng Fu was supported by Pierre and Christine Lamond Fellowship. We appreciate the help from HESSIAN and ARX.

\bibliographystyle{IEEEtran}
\bibliography{root}

@article{guo2025demospeedup,
  title={DemoSpeedup: Accelerating Visuomotor Policies via Entropy-Guided Demonstration Acceleration},
  author={Guo, Lingxiao and Xue, Zhengrong and Xu, Zijing and Xu, Huazhe},
  journal={arXiv preprint arXiv:2506.05064},
  year={2025}
}

@article{kim2026espada,
  title={Espada: Execution speedup via semantics aware demonstration data downsampling for imitation learning},
  author={Kim, Byungju and Pahk, Jinu and Lee, Chungwoo and Kim, Jaejoon and Lee, Jangha and Kim, Theo Taeyeong and Shim, Kyuhwan and Lee, Jun Ki and Zhang, Byoung-Tak},
  journal={IEEE Robotics and Automation Letters},
  year={2026},
  publisher={IEEE}
}

@article{han2026b,
  title={B-spline Policy: Accelerating Manipulation Policies via B-spline Action Representations},
  author={Han, Xiaoshen and Xiong, Haoyu and Chen, Haonan and Liu, Chaoqi and Torralba, Antonio and Zhu, Yuke and Du, Yilun},
  journal={arXiv preprint arXiv:2607.09648},
  year={2026}
}

@inproceedings{fu2024mobile,
  author    = {Fu, Zipeng and Zhao, Tony Z. and Finn, Chelsea},
  title     = {Mobile ALOHA: Learning Bimanual Mobile Manipulation with Low-Cost Whole-Body Teleoperation},
  booktitle = {{CoRL}},
  year      = {2024},
}

@inproceedings{fu2024humanplus,
  author    = {Fu, Zipeng and Zhao, Qingqing and Wu, Qi and Wetzstein, Gordon and Finn, Chelsea},
  title     = {HumanPlus: Humanoid Shadowing and Imitation from Humans},
  booktitle = {{CoRL}},
  year      = {2024},
}

@article{xu2025dexumi,
  title={DexUMI: Using Human Hand as the Universal Manipulation Interface for Dexterous Manipulation},
  author={Xu, Mengda and Zhang, Han and Hou, Yifan and Xu, Zhenjia and Fan, Linxi and Veloso, Manuela and Song, Shuran},
  journal={arXiv preprint arXiv:2505.21864},
  year={2025}
}

@article{ravi2024sam2,
  title={SAM 2: Segment Anything in Images and Videos},
  author={Ravi, Nikhila and Gabeur, Valentin and Hu, Yuan-Ting and Hu, Ronghang and Ryali, Chaitanya and Ma, Tengyu and Khedr, Haitham and R{\"a}dle, Roman and Rolland, Chloe and Gustafson, Laura and Mintun, Eric and Pan, Junting and Alwala, Kalyan Vasudev and Carion, Nicolas and Wu, Chao-Yuan and Girshick, Ross and Doll{\'a}r, Piotr and Feichtenhofer, Christoph},
  journal={arXiv preprint arXiv:2408.00714},
  year={2024}
}

@inproceedings{zhou2023propainter,
   title={{ProPainter}: Improving Propagation and Transformer for Video Inpainting},
   author={Zhou, Shangchen and Li, Chongyi and Chan, Kelvin C.K and Loy, Chen Change},
   booktitle={Proceedings of IEEE International Conference on Computer Vision (ICCV)},
   year={2023}
}

@article{purushottam2023dynamic,
  title={Dynamic mobile manipulation via whole-body bilateral teleoperation of a wheeled humanoid},
  author={Purushottam, Amartya and Xu, Christopher and Jung, Yeongtae and Ramos, Joao},
  journal={IEEE Robotics and Automation Letters},
  year={2023},
  publisher={IEEE}
}

@inproceedings{seo2023deep,
  title={Deep Imitation Learning for Humanoid Loco-manipulation through Human Teleoperation},
  author={Seo, Mingyo and Han, Steve and Sim, Kyutae and Bang, Seung Hyeon and Gonzalez, Carlos and Sentis, Luis and Zhu, Yuke},
  booktitle={2023 Humanoids},
  year={2023},
}

@article{he2024learning,
  title={Learning Human-to-Humanoid Real-Time Whole-Body Teleoperation},
  author={He, Tairan and Luo, Zhengyi and Xiao, Wenli and Zhang, Chong and Kitani, Kris and Liu, Changliu and Shi, Guanya},
  journal={arXiv preprint arXiv:2403.04436},
  year={2024}
}

@article{sivakumar2022robotic,
  title={Robotic telekinesis: Learning a robotic hand imitator by watching humans on youtube},
  author={Sivakumar, Aravind and Shaw, Kenneth and Pathak, Deepak},
  journal={arXiv preprint arXiv:2202.10448},
  year={2022}
}

@inproceedings{chi2024universal,
	title={Universal Manipulation Interface: In-The-Wild Robot Teaching Without In-The-Wild Robots},
	author={Chi, Cheng and Xu, Zhenjia and Pan, Chuer and Cousineau, Eric and Burchfiel, Benjamin and Feng, Siyuan and Tedrake, Russ and Song, Shuran},
	booktitle={Proceedings of Robotics: Science and Systems (RSS)},
	year={2024}
}

@article{wang2024dexcap,
  title = {DexCap: Scalable and Portable Mocap Data Collection System for Dexterous Manipulation},
  author = {Wang, Chen and Shi, Haochen and Wang, Weizhuo and Zhang, Ruohan and Fei-Fei, Li and Liu, C. Karen},
  journal = {arXiv preprint arXiv:2403.07788},
  year = {2024}
}

@article{zhao2023learning,
  title={Learning fine-grained bimanual manipulation with low-cost hardware},
  author={Zhao, Tony Z and Kumar, Vikash and Levine, Sergey and Finn, Chelsea},
  journal={RSS},
  year={2023}
}

@article{ito2022efficient,
  title={Efficient multitask learning with an embodied predictive model for door opening and entry with whole-body control},
  author={Ito, Hiroshi and Yamamoto, Kenjiro and Mori, Hiroki and Ogata, Tetsuya},
  journal={Science Robotics},
  year={2022},
}

@article{cheng2024tv,
    title={Open-TeleVision: Teleoperation with Immersive Active Visual Feedback},
    author={Cheng, Xuxin and Li, Jialong and Yang, Shiqi and Yang, Ge and Wang, Xiaolong},
    journal={arXiv preprint arXiv:2407.01512},
    year={2024}
}

@String(ICCV= {Int. Conf. Comput. Vis.})

@String(ICCV  = {ICCV})

@incollection{billard2008robot,
  title={Robot programming by demonstration},
  author={Billard, Aude and Calinon, Sylvain and Dillmann, Ruediger and Schaal, Stefan},
  booktitle={Springer handbook of robotics},
  pages={1371--1394},
  year={2008},
  publisher={Springer}
}

@article{xiong2024adaptive,
  title={Adaptive mobile manipulation for articulated objects in the open world},
  author={Xiong, Haoyu and Mendonca, Russell and Shaw, Kenneth and Pathak, Deepak},
  journal={arXiv preprint arXiv:2401.14403},
  year={2024}
}

@inproceedings{zhu2023viola,
  title={Viola: Imitation learning for vision-based manipulation with object proposal priors},
  author={Zhu, Yifeng and Joshi, Abhishek and Stone, Peter and Zhu, Yuke},
  booktitle={CoRL},
  pages={1199--1210},
  year={2023},
  organization={PMLR}
}

@inproceedings{ross2013learning,
  title={Learning monocular reactive uav control in cluttered natural environments},
  author={Ross, St{\'e}phane and Melik-Barkhudarov, Narek and Shankar, Kumar Shaurya and Wendel, Andreas and Dey, Debadeepta and Bagnell, J Andrew and Hebert, Martial},
  booktitle={2013 ICRA},
  year={2013}
}

@inproceedings{mandlekar2018roboturk,
  title={Roboturk: A crowdsourcing platform for robotic skill learning through imitation},
  author={Mandlekar, Ajay and Zhu, Yuke and Garg, Animesh and Booher, Jonathan and Spero, Max and Tung, Albert and Gao, Julian and Emmons, John and Gupta, Anchit and Orbay, Emre and others},
  booktitle={CoRL},
  pages={879--893},
  year={2018},
  organization={PMLR}
}

@article{park2024dexhub,
  title={Dexhub and dart: Towards internet scale robot data collection},
  author={Park, Younghyo and Bhatia, Jagdeep Singh and Ankile, Lars and Agrawal, Pulkit},
  journal={arXiv preprint arXiv:2411.02214},
  year={2024}
}

@inproceedings{fang2024airexo,
  title={Airexo: Low-cost exoskeletons for learning whole-arm manipulation in the wild},
  author={Fang, Hongjie and Fang, Hao-Shu and Wang, Yiming and Ren, Jieji and Chen, Jingjing and Zhang, Ruo and Wang, Weiming and Lu, Cewu},
  booktitle={2024 ICRA},
  year={2024},
}

@inproceedings{wu2024gello,
  title={Gello: A general, low-cost, and intuitive teleoperation framework for robot manipulators},
  author={Wu, Philipp and Shentu, Yide and Yi, Zhongke and Lin, Xingyu and Abbeel, Pieter},
  booktitle={2024 IROS},
  year={2024},
}

@inproceedings{buamanee2024bi,
  title={Bi-act: Bilateral control-based imitation learning via action chunking with transformer},
  author={Buamanee, Thanpimon and Kobayashi, Masato and Uranishi, Yuki and Takemura, Haruo},
  booktitle={2024 AIM},
  year={2024},
}

@inproceedings{etukuru2025robot,
  title={Robot utility models: General policies for zero-shot deployment in new environments},
  author={Etukuru, Haritheja and Naka, Norihito and Hu, Zijin and Lee, Seungjae and Mehu, Julian and Edsinger, Aaron and Paxton, Chris and Chintala, Soumith and Pinto, Lerrel and Shafiullah, Nur Muhammad Mahi},
  booktitle={2025 ICRA},
  year={2025},
}

@inproceedings{kareer2025egomimic,
  title={Egomimic: Scaling imitation learning via egocentric video},
  author={Kareer, Simar and Patel, Dhruv and Punamiya, Ryan and Mathur, Pranay and Cheng, Shuo and Wang, Chen and Hoffman, Judy and Xu, Danfei},
  booktitle={2025 ICRA},
  year={2025},
}

@article{lepert2025phantom,
  title={Phantom: Training robots without robots using only human videos},
  author={Lepert, Marion and Fang, Jiaying and Bohg, Jeannette},
  journal={arXiv preprint arXiv:2503.00779},
  year={2025}
}

@article{zhu2025emma,
  title={EMMA: Scaling Mobile Manipulation via Egocentric Human Data},
  author={Zhu, Lawrence Y and Kuppili, Pranav and Punamiya, Ryan and Aphiwetsa, Patcharapong and Patel, Dhruv and Kareer, Simar and Ha, Sehoon and Xu, Danfei},
  journal={arXiv preprint arXiv:2509.04443},
  year={2025}
}

@inproceedings{zhang2018deep,
  title={Deep imitation learning for complex manipulation tasks from virtual reality teleoperation},
  author={Zhang, Tianhao and McCarthy, Zoe and Jow, Owen and Lee, Dennis and Chen, Xi and Goldberg, Ken and Abbeel, Pieter},
  booktitle={2018 ICRA},
  year={2018},
}

@inproceedings{handa2020dexpilot,
  title={Dexpilot: Vision-based teleoperation of dexterous robotic hand-arm system},
  author={Handa, Ankur and Van Wyk, Karl and Yang, Wei and Liang, Jacky and Chao, Yu-Wei and Wan, Qian and Birchfield, Stan and Ratliff, Nathan and Fox, Dieter},
  booktitle={2020 ICRA},
  year={2020},
}

@inproceedings{finn2016guided,
  title={Guided cost learning: Deep inverse optimal control via policy optimization},
  author={Finn, Chelsea and Levine, Sergey and Abbeel, Pieter},
  booktitle={International conference on machine learning},
  year={2016},
}

@article{xie2019improvisation,
  title={Improvisation through physical understanding: Using novel objects as tools with visual foresight},
  author={Xie, Annie and Ebert, Frederik and Levine, Sergey and Finn, Chelsea},
  journal={arXiv preprint arXiv:1904.05538},
  year={2019}
}

@inproceedings{akgun2012trajectories,
  title={Trajectories and keyframes for kinesthetic teaching: A human-robot interaction perspective},
  author={Akgun, Baris and Cakmak, Maya and Yoo, Jae Wook and Thomaz, Andrea Lockerd},
  booktitle={Proceedings of the seventh annual ACM/IEEE international conference on Human-Robot Interaction},
  year={2012}
}

@inproceedings{kormushev2011upper,
  title={Upper-body kinesthetic teaching of a free-standing humanoid robot},
  author={Kormushev, Petar and Nenchev, Dragomir N and Calinon, Sylvain and Caldwell, Darwin G},
  booktitle={2011 ICRA},
  year={2011},
}

@article{kormushev2011imitation,
  title={Imitation learning of positional and force skills demonstrated via kinesthetic teaching and haptic input},
  author={Kormushev, Petar and Calinon, Sylvain and Caldwell, Darwin G},
  journal={Advanced Robotics},
  year={2011},
}

@inproceedings{weiss2009teaching,
  title={Teaching a humanoid: A user study on learning by demonstration with hoap-3},
  author={Weiss, Astrid and Igelsbock, Judith and Calinon, Sylvain and Billard, Aude and Tscheligi, Manfred},
  booktitle={RO-MAN 2009-The 18th IEEE International Symposium on Robot and Human Interactive Communication},
  year={2009},
  organization={IEEE}
}

@inproceedings{ben2009kinesthetic,
  title={Kinesthetic bootstrapping: Teaching motor skills to humanoid robots through physical interaction},
  author={Ben Amor, Heni and Berger, Erik and Vogt, David and Jung, Bernhard},
  booktitle={Annual conference on artificial intelligence},
  year={2009},
  organization={Springer}
}

@inproceedings{ichnowski2022gomp,
  title={Gomp-fit: Grasp-optimized motion planning for fast inertial transport},
  author={Ichnowski, Jeffrey and Avigal, Yahav and Liu, Yi and Goldberg, Ken},
  booktitle={2022 iICRA},
  pages={5255--5261},
  year={2022},
  organization={IEEE}
}

@article{kim2014catching,
  title={Catching objects in flight},
  author={Kim, Seungsu and Shukla, Ashwini and Billard, Aude},
  journal={IEEE Transactions on Robotics},
  year={2014},
}

@inproceedings{senoo2006ball,
  title={Ball control in high-speed batting motion using hybrid trajectory generator},
  author={Senoo, Taku and Namiki, Akio and Ishikawa, Masatoshi},
  booktitle={Proceedings 2006 ICRA.},
  year={2006},
}

@inproceedings{dambrosio2025achieving,
  title={Achieving human level competitive robot table tennis},
  author={DAmbrosio, David B and Abeyruwan, Saminda and Graesser, Laura and Iscen, Atil and Amor, Heni Ben and Bewley, Alex and Reed, Barney J and Reymann, Krista and Takayama, Leila and Tassa, Yuval and others},
  booktitle={2025 ICRA},
  year={2025},
}

@article{buchler2022learning,
  title={Learning to play table tennis from scratch using muscular robots},
  author={B{\"u}chler, Dieter and Guist, Simon and Calandra, Roberto and Berenz, Vincent and Sch{\"o}lkopf, Bernhard and Peters, Jan},
  journal={IEEE Transactions on Robotics},
  year={2022},
}

@article{mulling2013learning,
  title={Learning to select and generalize striking movements in robot table tennis},
  author={M{\"u}lling, Katharina and Kober, Jens and Kroemer, Oliver and Peters, Jan},
  journal={The International Journal of Robotics Research},
  year={2013},
}

@inproceedings{zhu2018towards,
  title={Towards high level skill learning: Learn to return table tennis ball using monte-carlo based policy gradient method},
  author={Zhu, Yifeng and Zhao, Yongsheng and Jin, Lisen and Wu, Jun and Xiong, Rong},
  booktitle={2018 IEEE international conference on real-time computing and robotics (RCAR)},
  year={2018},
}

@article{liu2013racket,
  title={Racket control for a table tennis robot to return a ball},
  author={Liu, Chunfang and Hayakawa, Yoshikazu and Nakashima, Akira},
  journal={SICE Journal of Control, Measurement, and System Integration},
  year={2013},
}

@inproceedings{zhang2021robots,
  title={Robots of the lost arc: Self-supervised learning to dynamically manipulate fixed-endpoint cables},
  author={Zhang, Harry and Ichnowski, Jeffrey and Seita, Daniel and Wang, Jonathan and Huang, Huang and Goldberg, Ken},
  booktitle={2021 ICRA},
  pages={4560--4567},
  year={2021},
  organization={IEEE}
}

@article{chi2024iterative,
  title={Iterative residual policy: for goal-conditioned dynamic manipulation of deformable objects},
  author={Chi, Cheng and Burchfiel, Benjamin and Cousineau, Eric and Feng, Siyuan and Song, Shuran},
  journal={The International Journal of Robotics Research},
  year={2024},
}

@article{zeng2020tossingbot,
  title={Tossingbot: Learning to throw arbitrary objects with residual physics},
  author={Zeng, Andy and Song, Shuran and Lee, Johnny and Rodriguez, Alberto and Funkhouser, Thomas},
  journal={IEEE Transactions on Robotics},
  year={2020},
}

@inproceedings{ha2022flingbot,
  title={Flingbot: The unreasonable effectiveness of dynamic manipulation for cloth unfolding},
  author={Ha, Huy and Song, Shuran},
  booktitle={CoRL},
  year={2022},
}

@inproceedings{yuan2025speedtuning,
  title={Speedtuning: Speeding Up Policy Execution with Lightweight Reinforcement Learning},
  author={Yuan, David D and Zhao, Tony Z and Burns, Kaylee and Finn, Chelsea},
  booktitle={2025 ICRA},
  year={2025},
}

@article{jiang2025behavior,
  title={Behavior robot suite: Streamlining real-world whole-body manipulation for everyday household activities},
  author={Jiang, Yunfan and Zhang, Ruohan and Wong, Josiah and Wang, Chen and Ze, Yanjie and Yin, Hang and Gokmen, Cem and Song, Shuran and Wu, Jiajun and Fei-Fei, Li},
  journal={arXiv preprint arXiv:2503.05652},
  year={2025}
}

@article{masuya2025variable,
  title={Variable-speed teaching--playback as real-world data augmentation for imitation learning},
  author={Masuya, Nozomu and Sato, Hiroshi and Yamane, Koki and Kusume, Takuya and Sakaino, Sho and Tsuji, Toshiaki},
  journal={Advanced Robotics},
  year={2025},
}

@article{arachchige2025sail,
  title={SAIL: Faster-than-Demonstration Execution of Imitation Learning Policies},
  author={Arachchige, Nadun Ranawaka and Chen, Zhenyang and Jung, Wonsuhk and Shin, Woo Chul and Bansal, Rohan and Barroso, Pierre and He, Yu Hang and Lin, Yingyang Celine and Joffe, Benjamin and Kousik, Shreyas and others},
  journal={arXiv preprint arXiv:2506.11948},
  year={2025}
}

@techreport{jenness1975master,
  title={Master--slave manipulators and remote maintenance at the Oak Ridge National Laboratory},
  author={Jenness, RG and Wicker, CD},
  year={1975},
  institution={Oak Ridge National Lab., Tenn.(USA)}
}

@book{lynch2017modern,
  title={Modern robotics},
  author={Lynch, Kevin M and Park, Frank C},
  year={2017},
  publisher={Cambridge University Press}
}

@article{black2025real,
  title={Real-Time Execution of Action Chunking Flow Policies},
  author={Black, Kevin and Galliker, Manuel Y and Levine, Sergey},
  journal={arXiv preprint arXiv:2506.07339},
  year={2025}
}

@article{zhu2020robosuite,
  title={robosuite: A Modular Simulation Framework and Benchmark for Robot Learning},
  author={Zhu, Yuke and Wong, Josiah and Mandlekar, Ajay and Mart{\'i}n-Mart{\'i}n, Roberto},
  journal={arXiv preprint arXiv:2009.12293},
  year={2020}
}

\clearpage
\appendices
\section{Tracking Error During Policy Rollouts}
\begin{figure*}[t]
    \centering
    \includegraphics[width=0.8\linewidth]{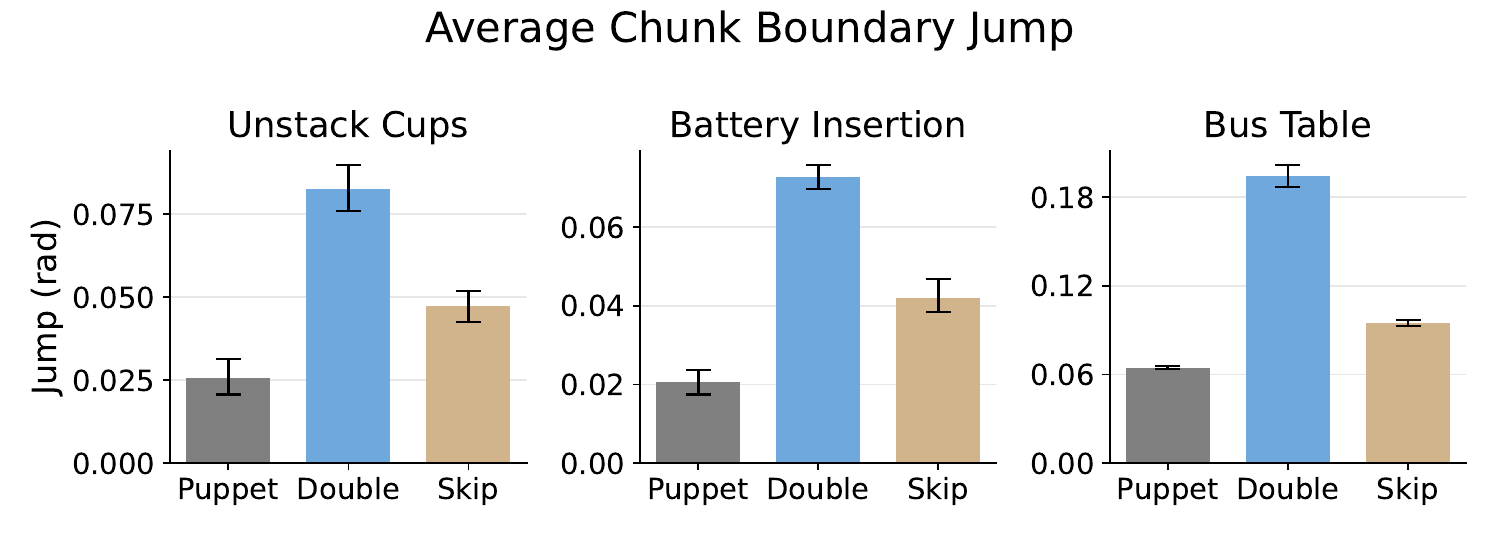}
    \caption{\textbf{\textit{Average Chunk Boundary Jump.}} Average change in commanded joint positions at chunk boundaries for Puppeteering (Puppet), Puppet Double Freq (Double) and Puppet Skip Actions (Skip). Bars: mean over 20 rollouts per condition (19 for Puppeteering on \textbf{\textit{Bus Table}})}
    \label{fig:chunk-jump}
\end{figure*}

\begin{figure}[t]
    \centering
    \includegraphics[width=\linewidth]{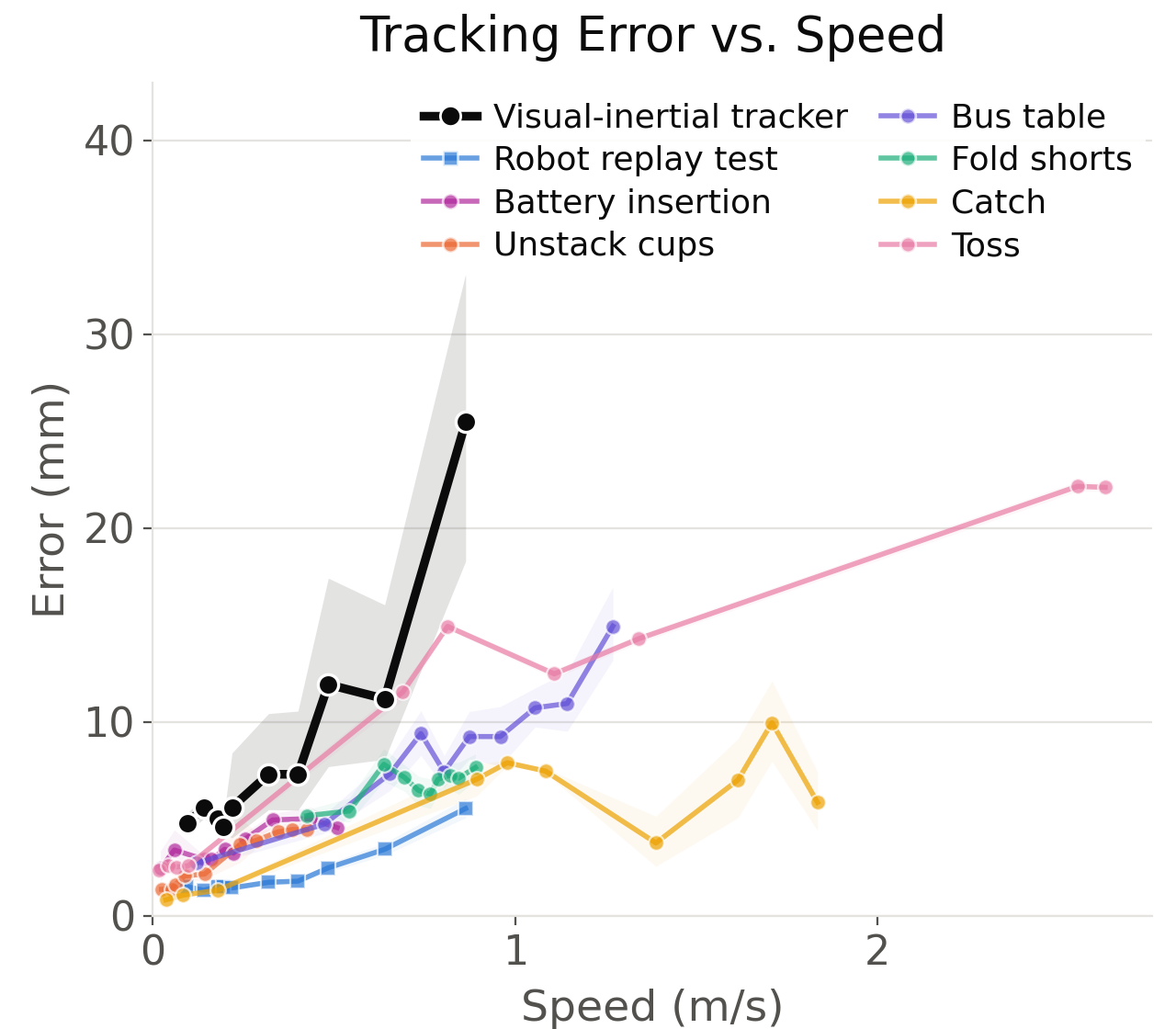}
    \caption{\textbf{\textit{Tracking Error During Policy Rollouts.}} Average position error in 1\,s windows, binned by end-effector speed (shaded: 95\% bootstrap CI). Black: a handheld visual-inertial tracker mounted on the robot wrist, against encoder ground truth. Colored: commanded vs.\ achieved end-effector position on our arms, during a tracking-test replay and policy rollouts of each task. Visual-inertial error grows quickly with speed and exceeds the robot's own tracking error at every speed it was tested.}
    \label{fig:rollout-tracking}
\end{figure}

\begin{figure}[t]
    \centering
    \includegraphics[width=0.85\linewidth]{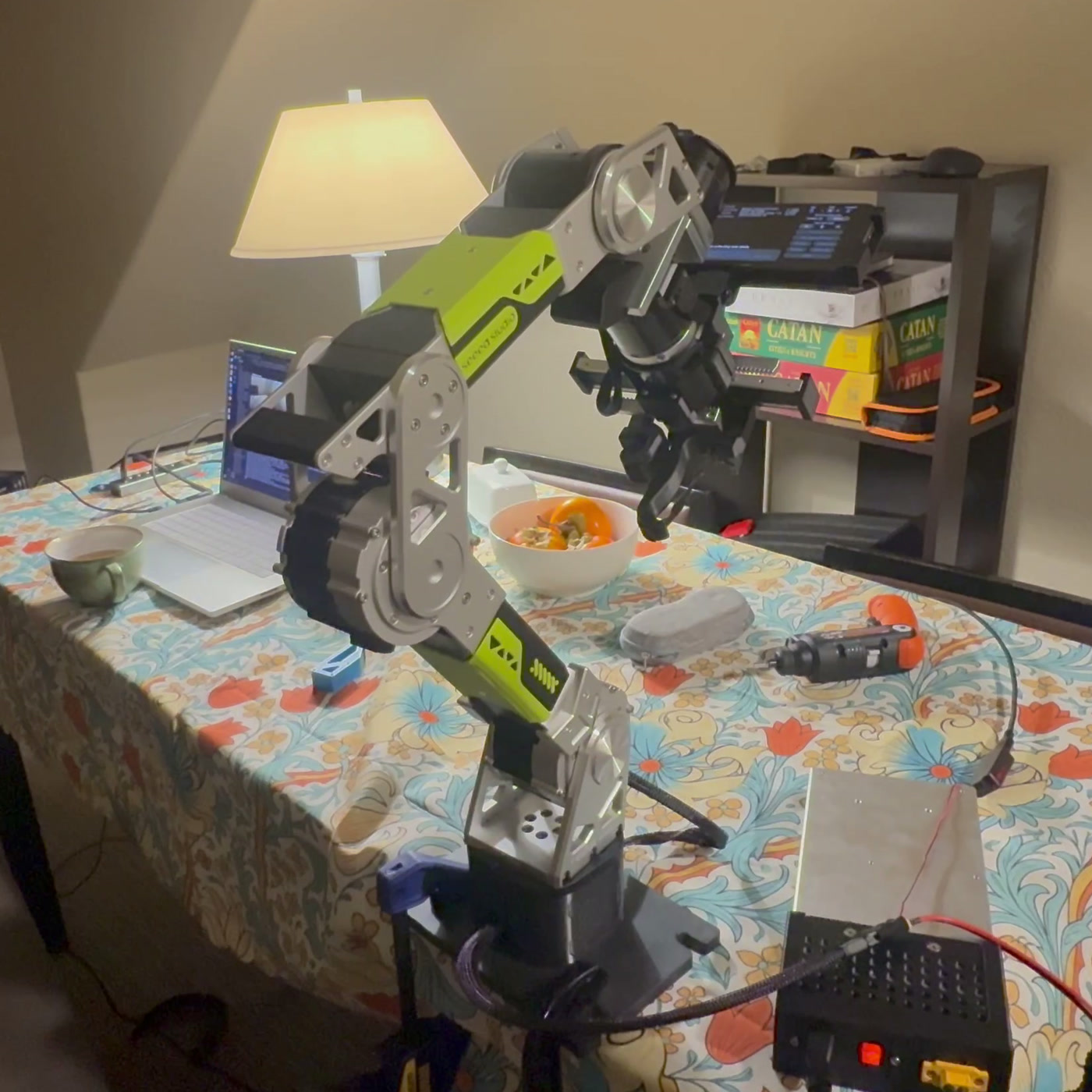}
    \caption{\textbf{\textit{Visual-Inertial Setup.}} We add features beneficial to visual-inertial tracking, including a patterned tablecloth, bright lighting, 3-dimensional objects, and background clutter for maximum tracking performance.}
    \label{fig:vio-setup}
\end{figure}

\label{app:rollout-tracking}
Fig.~\ref{fig:rollout-tracking} extends Fig.~\ref{fig:vio-tracking} with the commanded vs.\ achieved end-effector error of our arms during policy rollouts of each task.
Fig.~\ref{fig:vio-setup} shows the visual-inertial tracking setup.

\section{Visual-Inertial Tracking Experiment}
\label{app:vio-setup}
\noindent\textbf{Hardware.} We mount an iPhone~15~Pro running the iPhUMI app on the wrist of a 6-DoF Seeed Studio reBot B601-RS arm with RobStride actuators, replacing the arm's last link with the iPhUMI phone holder and hollow gripper fingers (Fig.~\ref{fig:vio-setup}). iPhUMI records the phone camera's ARKit visual-inertial pose at about 60\,Hz. It cannot stream poses live, so each recording is exported from the phone and aligned with the robot data offline. The phone stays seated in the holder throughout, and the transform $X$ from the end effector to the phone camera is taken from the holder's CAD model.

\noindent\textbf{Ground truth.} The arm records its joint encoders, and forward kinematics gives the end-effector pose $A_i$ at each sample. ARKit's world frame is gravity-aligned, but its origin and heading are set wherever tracking starts, so it differs between recordings. For each take we fit one rigid transform $Y$ from the ARKit world frame to the robot base by least squares over the whole take, and compare $A_i$ with the tracker's estimate of the end-effector pose
\[
\hat{A}_i = Y B_i X^{-1},
\]
where $B_i$ is the ARKit camera pose.

\noindent\textbf{Time alignment.} The host and the phone are NTP-synchronized, and robot timestamps are mapped to Unix time through clock anchors recorded during each session. The remaining offset between the two streams, including pose latency, is found per take by cross-correlating the end effector's angular speed from both sources, which does not depend on $X$ or $Y$; the peak correlation is at least 0.97 for every take. Tracker poses are then interpolated to the robot samples (linearly in position, spherically in rotation), and robot samples that fall in a gap of the tracker log are excluded rather than bridged.

\noindent\textbf{Trajectories.} An operator hand-guides the arm through four trajectories at 100\,Hz while the phone records: a general motion, a free-space motion, a high-frequency motion and a motion with contact. The arm then replays each trajectory from its recorded joint positions at 0.25$\times$, 0.5$\times$, 0.75$\times$ and 1$\times$ speed, again with the phone recording. Including the hand-guided recordings, this gives 48 takes and 1,269 one-second windows, with end-effector speeds up to 1.2\,m/s.

\noindent\textbf{Error metric.} Each take is split into 1\,s windows, and each window reports its top end-effector speed and the average end-effector position error. Windows with missing tracker samples are excluded. The curves in Fig.~\ref{fig:vio-tracking} and Fig.~\ref{fig:rollout-tracking} are the mean over ten equal-count speed bins, with 95\% bootstrap confidence intervals. For the robot's own tracking error during the replays, we compare the commanded and achieved end-effector positions after removing one constant time lag per take, chosen to minimize the RMS error.

\noindent\textbf{Recording Failures.} On the 1$\times$ speed takes for each trajectory, SLAM failures are rejected where the end effector is detected moving at more than 7.0 m/s by the iPhUMI kit. Due to this, each take at 1$\times$ speed on average took 3-4 tries to record without failure, making it impractical to test iPhUMI tracking on even faster trajectories.

\section{Puppeteering Baseline Setup}
\label{app:aloha2}
\noindent\textbf{Data collection.} For the puppeteering comparisons in Table~\ref{tab:puppeteering}, an operator collects demonstrations by teleoperating an ALOHA~2 with its leader arms. Each task has 50 demonstrations recorded at 50\,Hz, with joint positions of both arms and grippers (14 dimensions) and images from a top camera and two wrist cameras at $640\times480$. Table~\ref{tab:app-aloha2} lists the datasets.

\noindent\textbf{Policy.} We train one ACT policy per task with the same architecture as ours (Table~\ref{tab:app-train}): ResNet-18 image backbone, no CVAE encoder, hidden dimension 512, feedforward dimension 3200, 4 encoder and 7 decoder layers with 8 heads, and a chunk size of 50 actions (1\,s at 50\,Hz). Policies take the three images at $320\times180$ and the joint positions, and are trained with batch size 256, learning rate $2\times10^{-4}$, backbone learning rate $10^{-5}$ and data augmentation.

\noindent\textbf{Deployment.} \textit{Puppeteering} executes each chunk at the 50\,Hz collection rate and queries a new chunk every 1\,s. \textit{Puppet Double Freq} executes the same chunks at 100\,Hz and queries every 0.5\,s. \textit{Puppet Skip Actions} keeps the 50\,Hz controller but executes every other action of each chunk, also querying every 0.5\,s. Both sped-up variants therefore traverse each chunk in half the time of the collected data.

\begin{table}[h]
\centering
\footnotesize
\setlength{\tabcolsep}{4pt}
\begin{tabular}{@{}lccc@{}}
\toprule
Task & Demos & Length & Training steps \\
\midrule
Unstack Cups      & 50 & 12\,s & 200k \\
Battery Insertion & 50 & 20\,s & 200k \\
Bus Table         & 50 & 18\,s & 200k \\
\bottomrule
\end{tabular}
\caption{\textbf{\textit{Puppeteering Data.}} ALOHA~2 teleoperation datasets used for the puppeteering baselines, recorded at 50\,Hz; Length is the recorded episode length.}
\label{tab:app-aloha2}
\end{table}

\section{Chunk Boundary Jumps}
\label{app:chunk-jump}
Fig.~\ref{fig:chunk-jump} measures the discontinuities between action chunks in the speed comparisons of Table~\ref{tab:puppeteering}. At each policy update, we take the change in commanded joint positions between the last action of the previous chunk and the first action of the new one, as the Euclidean norm over the 12 arm joints (grippers excluded). Each rollout contributes the average jump over its chunk boundaries, and every rollout is included, failures as well. Doubling the controller frequency makes the jumps about 3--3.5$\times$ those of puppeteering, while skipping every other action makes them 1.5--2$\times$.

\section{Policy Training Details}
\label{app:training}
We train ACT~\cite{zhao2023learning} policies on the demonstrations. Table~\ref{tab:app-data} lists the datasets and Table~\ref{tab:app-train} the ACT configuration. Images are recorded at $360\times640$ and downsampled to $180\times320$ for the policy, and training augments them with random crops covering 95\% of each side, resized back to the input resolution. Policies predict absolute joint positions and gripper commands, are conditioned on joint positions only (no velocities), and are trained with seed 0 on a single GPU (NVIDIA L40S, A40 or RTX 4090), holding out 1\% of the episodes for validation.

\begin{table}[h]
\centering
\footnotesize
\setlength{\tabcolsep}{3pt}
\resizebox{\linewidth}{!}{%
\begin{tabular}{@{}lccccc@{}}
\toprule
Task & Arms & Cameras & Rate & Demos & Length \\
\midrule
Unstack Cups      & ARX, 2 & L, R, Top & 50\,Hz  & 50  & 4\,s \\
Battery Insertion & ARX, 2 & L, R, Top & 50\,Hz  & 50  & 7\,s \\
Bus Table         & ARX, 2 & L, R, Top & 50\,Hz  & 50  & 6\,s \\
Fold Shorts       & ARX, 2 & L, R, Top & 50\,Hz  & 50  & 5\,s \\
Toss \& Catch     & HESSIAN, 2 & L, R, Top & 100/200\,Hz & 50  & 2\,s \\
\bottomrule
\end{tabular}}
\caption{\textbf{\textit{Training Data.}} Demonstrations per task, recorded by kinesthetic teaching. L and R are the wrist cameras; Length is the task duration. Toss \& Catch records the tossing arm at 100\,Hz and the catching arm at 200\,Hz.}
\label{tab:app-data}
\end{table}

\begin{table}[h]
\centering
\footnotesize
\begin{tabular}{@{}ll@{}}
\toprule
Policy                    & ACT, no CVAE encoder \\
Image backbone            & ResNet-18 \\
Hidden / feedforward dim. & 512 / 3200 \\
Encoder / decoder layers  & 4 / 7 \\
Attention heads           & 8 \\
Chunk size                & 50 \\
Batch size                & 256 \\
Learning rate             & $2\times10^{-4}$ \\
Optimizer                 & AdamW, weight decay $10^{-4}$ \\
Training steps            & 200k \\
\bottomrule
\end{tabular}
\caption{\textbf{\textit{ACT Configuration.}}}
\label{tab:app-train}
\end{table}

\end{document}